 \documentclass[preprint,review,12pt]{elsarticle}

\usepackage{graphics}
\usepackage{graphicx}
\usepackage{epsfig}

\usepackage{amssymb}
\usepackage{amsthm}
\usepackage{amsmath}

\journal{Medical Image Analysis}

\begin{document}

\begin{frontmatter}



\title{MORPHOLOGICAL ANALYSIS OF THE LEFT VENTRICULAR ENDOCARDIAL SURFACE USING A BAG-OF-FEATURES DESCRIPTOR}


\author{Anirban Mukhopadhyay$^{\dagger}$, Zhen Qian$^{\ast}$, Suchendra M. Bhandarkar$^{\dagger}$, Tianming Liu$^{\dagger}$, Sarah Rinehart$^{\ast}$, Szilard Voros$^{\ddagger}$}

\address{$^{\dagger}$Department of Computer Science, The University of Georgia, Athens, GA, USA\\
$^{\ast}$Piedmont Heart Institute, Atlanta, GA, USA\\
$^{\ddagger}$Stony Brook University Medical Center, Stony Brook, NY, USA
}

\begin{abstract}

The limitations of conventional imaging techniques have hitherto precluded a thorough and formal investigation of the complex morphology of the left ventricular (LV) endocardial surface and its relation to the severity of Coronary Artery Disease (CAD). However, recent developments in high-resolution Multirow-Detector Computed Tomography (MDCT) scanner technology have enabled the imaging of the complex LV endocardial surface morphology in a single heart beat. Analysis of high-resolution Computed Tomography (CT) images from a 320-MDCT scanner allows for the study of the relationship between percent Diameter Stenosis (DS) of the major coronary arteries and localization of the cardiac segments affected by coronary arterial stenosis. In this paper a novel approach for the analysis of the non-rigid LV endocardial surface from MDCT images, using a combination of rigid body transformation-invariant shape descriptors and a more generalized isometry-invariant {\it Bag-of-Features} (BoF) descriptor, is proposed and implemented. The proposed approach is shown to be successful in identifying, localizing and quantifying the incidence and extent of CAD and thus, is seen to have a potentially significant clinical impact. Specifically, the association between the incidence and extent of CAD, determined via the percent DS measurements of the major coronary arteries, and the alterations in the endocardial surface morphology is formally quantified. The results of the proposed approach on 16 normal data sets, and 16 abnormal data sets exhibiting CAD with varying levels of severity, are presented. A multivariate regression test is employed to test the effectiveness of the proposed morphological analysis approach. Experiments performed on a strict leave-one-out basis are shown to exhibit a distinct pattern in terms of the correlation coefficient within the cardiac segments where the incidence of coronary arterial stenosis is localized.

\end{abstract}

\begin{keyword}

Ventricular endocardial surface,\ cardiovascular CT,\ non-rigid shape analysis,\ Shape-Index,\ Bag-of-Features.

\end{keyword}

\end{frontmatter}


\setcounter{secnumdepth}{5}

\section{Introduction} \label{intro}

Since {\it Coronary Artery Disease} (CAD) is a leading cause of morbidity and mortality worldwide~\cite{goo09}, techniques that increase the effectiveness and/or lower the costs of diagnostic or prognostic procedures associated with CAD are expected to have a significant clinical impact. CAD is caused by {\it atherosclerosis} or accumulation of lipoprotein plaque in the coronary arteries that supply blood to the {\it myocardium} or cardiac muscle tissue. Atherosclerotic plaques lead to the progressive narrowing or {\it stenosis} of the coronary arteries, resulting in reduced blood flow and consequently, reduced oxygen supply to the myocardium, a condition termed as {\it myocardial ischemia}. If untreated, myocardial ischemia may lead to irreversible necrosis of the myocardium wherein the healthy myocardium is increasingly replaced by scar tissue, thus compromising the cardiac function and resulting in congestive cardiac failure. Some vulnerable plaques may suddenly rupture resulting in coronary artery occlusion and cardiac arrest, potentially leading to  sudden death. 

X-Ray Coronary Angiography (XRA) is an invasive technique that is a clinically accepted standard for assessment of vascular morphology and the extent of vessel stenosis due to artherosclerotic plaque deposition. However, a {\it comprehensive} assessment of CAD necessitates a study of both, vascular morphology and cardiovascular function. Conventional cardiovascular functional assessment is performed via a stress-induced perfusion test that uses Magnetic Resonance (MR) or radionuclide Myocardial Perfusion Imaging (MPI). Since vascular morphology and cardiovascular function are imaged using separate modalities, the time and cost associated with a comprehensive assessment of CAD and the potential health risk to the patient associated with higher radiation doses are all significantly increased. 

\begin{figure}[t]
\begin{center}
   \includegraphics[width=80mm]{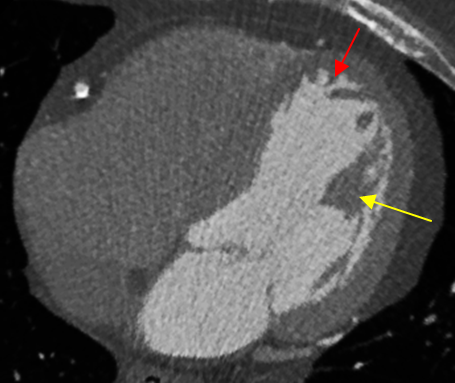}
\end{center}
   \caption{Left ventricular trabeculae carneae (pointed by the red arrow) and papillary muscles (pointed by the yellow arrow) can be clearly seen in the CTCA images acquired using a 320-MDCT scanner.}
\label{fig:intro_visual}
\end{figure}

Computed Tomography Coronary Angiography (CTCA) is a non-invasive imaging technique that allows for robust quantification of vascular morphology and has the potential for characterizing the atherosclerotic plaque composition~\cite{mukhopadhyay12}. When performed using a 320-Multirow-Detector Computed Tomography (MDCT) scanner, a CTCA scan can be performed within a single heart beat, yielding images with an isotropic spatial resolution of 0.5 mm in a volumetric fashion. The resulting images, in addition to providing vascular morphology information, are capable of providing significant details about the left ventricular (LV) endocardial surface structure. The CTCA images, backed by anatomical studies, have revealed that, instead of a simple and smooth surface, the LV endocardial surface is composed of a complex structure of {\it trabeculae carneae}, which are small muscular columns that arise naturally from the inner surface of the ventricles (Figure~\ref{fig:intro_visual}). Also revealed in the CTCA images are the {\it papillary} muscles which are attached to the cusps of the atrioventricular (i.e., the mitral and tricuspid) valves (Figure~\ref{fig:intro_visual}).

Alterations  in the LV trabeculation structure have been clinically observed to closely associate with some cardiovascular diseases, such as myocardial noncompaction disease~\cite{agmon99} and hypertrophy~\cite{goo09}. However, due to the limitations in the spatial resolution of conventional {\it in vivo} imaging techniques, very few research studies have been undertaken to formally investigate the LV trabeculation structure at a detailed level, and formally quantify the relationships between structural changes in LV trabeculation, changes in cardiovascular function and cardiovascular pathologies. If the analysis of the LV endocardial surface structure can be shown to provide significant insights into vascular morphology, cardiovascular function and the progression of CAD then it has the potential to provide significant additional or supplementary diagnostic value to the results of MPI, CTCA and XRA. In some cases, it could potentially reduce the need for invasive and/or stress-based testing procedures that are expensive, time consuming and pose a greater health risk to the patient, thus making CTCA imaging using MDCT scanners, a potential one-stop-shop technique for assessment of both, vascular morphology and cardiovascular function.   

\begin{figure*}
\begin{center}
 \includegraphics[width=130mm]{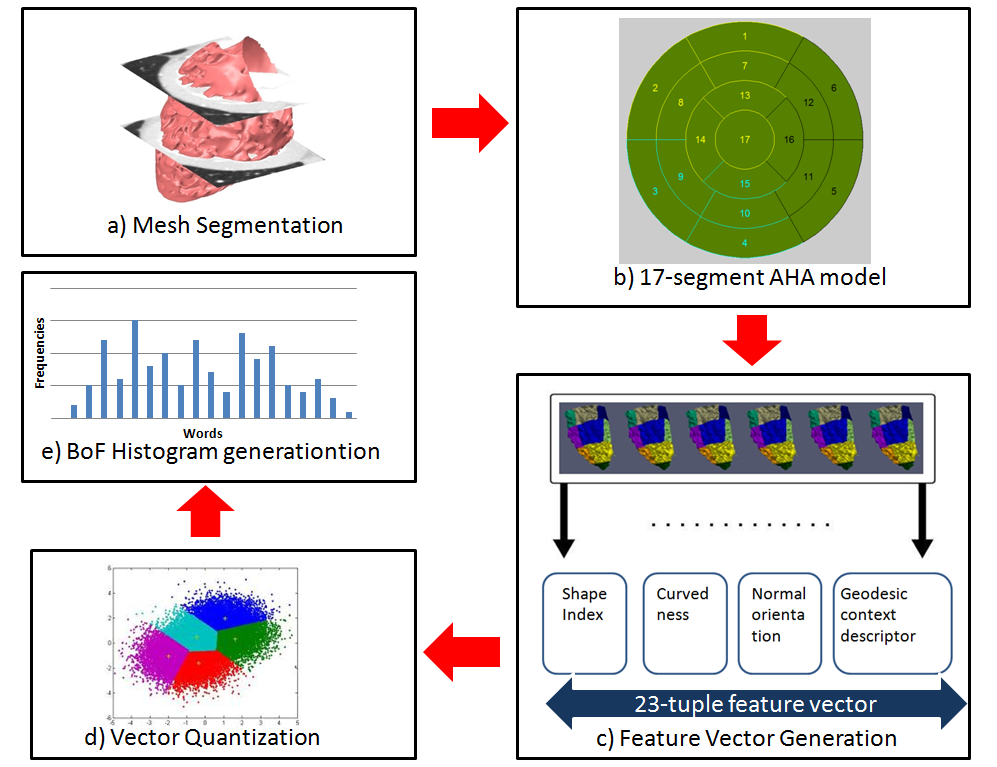}
\end{center}
   \caption{Illustration of the sequence of steps in the morphological analysis of the LV endocardial surface: (a) accurate mesh segmentation followed by (b) generation of a 17-segment LV surface model with demarcation of coronary arterial territories (red: LAD, green: LCX, blue: RCA), (c) feature vector generation and, (d, e) final generation of the BoF histogram via vector quantization ($k$-means clustering).}
\label{fig:concept_visual}
\end{figure*}

The sub-endocardial layer of the myocardium is the first and the most susceptible cardiac region in the development of CAD on account of its inherently higher oxygen consumption requirement and the restricted collateral blood flow resulting in reduced oxygen delivery during the progression of CAD. The presence of intermediate to severe CAD in patients is often clinically suggestive of varying degrees of myocardial ischemia in the corresponding regions of the myocardium. Furthermore, the affected ischemic myocardium loses contractibility and has a tendency to get stiffer and be pushed outward by the high LV blood pressure. These myocardial changes are reflected in the LV endocardial surface morphology, in that the structure of the trabeculae and the papillary muscles on the LV endocardial surface are clinically observed to be substantially altered~\cite{mukhopadhyay11}. This clinically hypothesized association between the LV endocardial surface morphology and the incidence and severity of CAD constitutes the central thesis of this paper.

The objective of this paper is to formally establish the clinically observed relationship between the incidence and severity of CAD, as quantified by the extent and location(s) of coronary arterial stenosis, and the structural alterations in the LV endocardial surface as observed in high-resolution MDCT images. The LV endocardial surface structure, specifically the structure of the LV endocardial trabeculae and papillary muscles, is formally characterized using 3D morphological features obtained via 3D shape analysis algorithms. To the best of our knowledge, this research is amongst the first to formally characterize the LV endocardial surface structure and quantify the clinically observed association between the alterations in the LV endocardial surface structure and the extent and location(s) of coronary arterial stenosis. The research described in this paper could potentially add significant value to the accuracy of diagnosis and effective management of CAD. It constitutes a first step towards formal investigation, quantification and verification of the clinically observed association between the LV endocardial surface morphology, cardiovascular function and vascular morphology which, in the long term, would aid in comprehensive assessment and understanding of the physiological mechanisms underlying the development and progression of CAD.

\section{Related Work}\label{related}

Our preliminary work~\cite{mukhopadhyay11} demonstrated the potential diagnostic value of characterization of the LV endocardial surface structure in assessing the incidence of CAD. Although it produced some significant and encouraging results, our preliminary work did exhibit shortcomings that stemmed primarily from the selection of shape descriptors used to characterize the LV endocardial surface. The two shape descriptors that were employed were based on the implicit assumption of rigidity of the LV endocardial surface as observed in the MDCT images. Consequently, the two shape descriptors were observed to have significant limitations in terms of their classification accuracy~\cite{mukhopadhyay11}. Although the MDCT image data were collected at a relatively steady phase of 75\% in the R-R cardiac cycle, the continuous LV motion and the human error involved in the data collection, severely tested the limits of the rigidity assumption. It was clear that a more robust shape descriptor was needed to characterize the LV endocardial surface, i.e., one that is invariant to isometric global shape deformation~\cite{memoli05,sun09}. 

Isometry-invariant shape descriptors have been the topic of recent research, especially in the context of content-based retrieval in shape databases~\cite{ovsjanikov09}, shape symmetry detection~\cite{mitra06}, dense surface correspondence determination~\cite{sun09} and surface registration~\cite{ovsjanikov10}. 
However, the problem in directly using existing shape descriptors developed for other applications, such as content-based image retrieval is that, these shape descriptors are designed primarily for differentiating between two distinct classes of objects (e.g., {\it humans} in different poses versus {\it dogs} in different poses). In contrast, our goal is to differentiate objects within the same class, e.g., to classify whether a particular LV segment is {\it “normal”} or {\it “diseased”}, based on its surface geometry and surface morphology. 
   
Deriving a feature-based representation of the contents of an image generally entails two stages of processing, {\it feature detection} and {\it feature description}~\cite{ovsjanikov09}. The goal of feature detection is to localize relatively stable points or regions within an image that possess significant information and can be repeatedly and reliably detected in transformed versions of the image. Different approaches are employed for feature detection based on the intrinsic nature of the features, i.e., point-based or region-based, and the desired scale of abstraction for the features. For example, in the case of the {\it Scale Invariant Feature Transform} (SIFT) feature detection technique, feature points of interest are localized by determining the local maxima of the discrete image Laplacian at varying scales of the Gaussian smoothing filter~\cite{lowe04}. Point-based feature selection techniques could either select all the points in the image or perform a dense sub-sampling of the set of feature points~\cite{mpeg04}. 

Region-based feature detection techniques typically rely on segmentation of the cardiac images using shape priors~\cite{comaniciu04,qian06}. The high-dimensional shape prior is projected into a low-dimensional subspace using dimensionality reduction techniques, e.g., variants of Principal Component Analysis (PCA), to constrain the shape variation. The detected features are tracked in a cardiac image sequence using a Kalman filter~\cite{comaniciu04} or particle filter~\cite{qian06} whereas the shape prior is adapted using a learning algorithm such as AdaBoost~\cite{qian06}. Other region-based feature detection techniques for cardiac images include model-based deformable templates~\cite{chalana96}, Markov random fields~\cite{mignotte01}, optical flow techniques~\cite{mailloux89} and combinations of the above~\cite{althoff00}. 

The goal of the feature description stage is to arrive at a representation of the local image information in the neighborhood of the detected feature. For example, the SIFT technique represents each feature point by a 128-dimensional descriptor vector that is invariant to scaling. A similar approach is taken in the {\it Speeded-Up Robust Feature} (SURF) transform approach~\cite{bay08}, where a 64-dimensional descriptor for each feature point is computed efficiently using integral images. In order to achieve a parsimonious representation, a feature vocabulary consisting of {\it visual words} is constructed by performing vector quantization in the feature descriptor space using a clustering technique such as the $k$-means clustering algorithm. After having performed vector quantization, the individual feature descriptors are replaced by indices in the vocabulary of visual words. The aggregation of feature descriptors to describe the overall shape is achieved by generating a frequency histogram of the visual words in the vocabulary, termed as a {\it Bag-of-Features} (BoF).  

One of the prominent implementations of the BoF concept is {\it Video Google}, a web-based application designed by Sivic and Zisserman~\cite{sivic03,sivic06} for object-based search in large image and video collections. In {\it Video Google} an image is represented by a collection of {\it visual words}. The visual words are generated using feature detectors and feature descriptors and indexed in a {\it visual vocabulary}. The BoF representation is constructed from the frequencies of the visual word occurrences in the image. Images with similar visual information are observed to have similar BoF representations. Thus, the comparison of BoF representations allows one to retrieve similar images. Likewise, two shapes are compared by comparing their respective visual word frequency histograms or BoF representations. This reduces the shape similarity problem to the problem of frequency histogram comparison. {\it Shape Google}, a technique for shape-based search in large image collections developed by Ovsjanikov et. al.~\cite{ovsjanikov09} and, the shape comparison approach proposed by Toldo et. al.~\cite{toldo10} are amongst the prominent works on BoF-based shape description and shape comparison, and the work described in this paper is influenced by both of these papers~\cite{ovsjanikov09,toldo10}. 

As an extension to the general BoF approach, we propose a novel BoF-based shape analysis approach designed specifically for cardiac image analysis, and for medical image analysis in general. The proposed BoF-based shape analysis approach is designed to quantify the relationship between the incidence, severity and localization of CAD and the structural alterations in the LV endocardial surface. While several works have proposed feature-based approaches for characterization of rigid 3D shapes, very few are capable of dealing with non-rigid 3D shape deformations~\cite{mitra05}. To the best of our knowledge, this is one of the first works that uses a BoF-based shape analysis approach for comparison of non-rigid deformable 3D shapes in the context of cardiovascular imaging in particular, and medical imaging in general.  

In this paper, a BoF-based approach is proposed to encapsulate the local and global geometry, local surface orientation and global contextual information for the LV endocardial surface. The proposed approach is shown to result in a robust feature vector for the purpose of morphological analysis of the LV endocardial surface. As shown in the experimental results, the proposed approach results in successful localization of coronary arterial stenosis which serves to strengthen the clinically observed relationship between the incidence and severity of CAD and morphological alterations in the LV endocardial surface. The sequence of steps in the proposed approach for morphological analysis of the LV endocardial surface is depicted in Figure~\ref{fig:concept_visual}.   

The remainder of the paper is organized as follows. In Section~\ref{sec:contribution}, the main contributions of the paper are outlined; in Section~\ref{sec:process}, the proposed LV surface segmentation and LV shape analysis procedures are detailed; and in Section~\ref{sec:result}, experimental results are presented. Finally, in Section~\ref{sec:discussion}, the paper is concluded with a brief discussion of the proposed approach and an outline of directions for future work.

\section{Contributions of the Paper}\label{sec:contribution}

\noindent 
The paper makes two primary contributions as described below:
\begin{enumerate}
\item The {\it Bag-of-Features} (BoF) framework for non-rigid shape analysis is adapted for the purpose of cardiac shape analysis which is an important problem in cardiac imaging in particular and medical imaging in general. It is also important to note that the proposed approach is sufficiently general to be applicable to other problems in medical imaging such as 3D shape analysis of the human brain and 3D shape-based search and retrieval in large medical databases. 

\item The paper proposes a geometric and machine learning-based model of the relationship between localized changes in the LV endocardial surface morphology and the incidence and extent of stenosis in specific coronary arteries. This constitutes an important initial step towards clinical understanding of the complex relationship between coronary arterial stenosis and its effect on the morphology of the LV endocardial surface. To the best of our knowledge, this paper represents one of the first attempts to model this complex clinical relationship in a mathematically structured manner.
\end{enumerate}

\section{MDCT Image Segmentation and LV Shape Analysis}\label{sec:process}

\subsection{LV endocardial surface segmentation and meshing}\label{subsec:LV_seg}

The {\it trabeculae carneae} along the LV endocardial surface can be broadly classified into three different morphological  types: (a) those that lie along the entire length of the LV wall forming prominent ridges; (b) those that are fixed at their extremities but free in the middle; and (c) those that connect the roots of the papillary muscles to the ventricular wall. These different {\it trabeculae} morphologies result in a complex LV endocardial surface topology. 

A 3D level set approach is employed to segment the LV endocardial surface while adapting to the topological changes caused by the complex trabeculation structure. A median filter-based denoising procedure is employed on the 3D MDCT data prior to segmentation in order to suppress noise while retaining the edges in the MDCT images. Unless mentioned otherwise, the size of the median filter is empirically set to $7\times7$ based on the MDCT data set. A level set-based segmentation procedure without reinitialization, as proposed by Li et al.~\cite{li05}, is applied to the median-filtered 3D image data set followed by the {\it Marching Cubes} procedure~\cite{lorensen87} to generate the surface meshes. The surface meshes are subsequently denoised via a mean face normal filtering procedure proposed by Zhang and Hamza~\cite{zhang07} to obtain the smooth shape of the LV myocardial surface.

\subsection{Data preparation}

\begin{figure}[t]
\begin{center}
   \includegraphics[width=80mm]{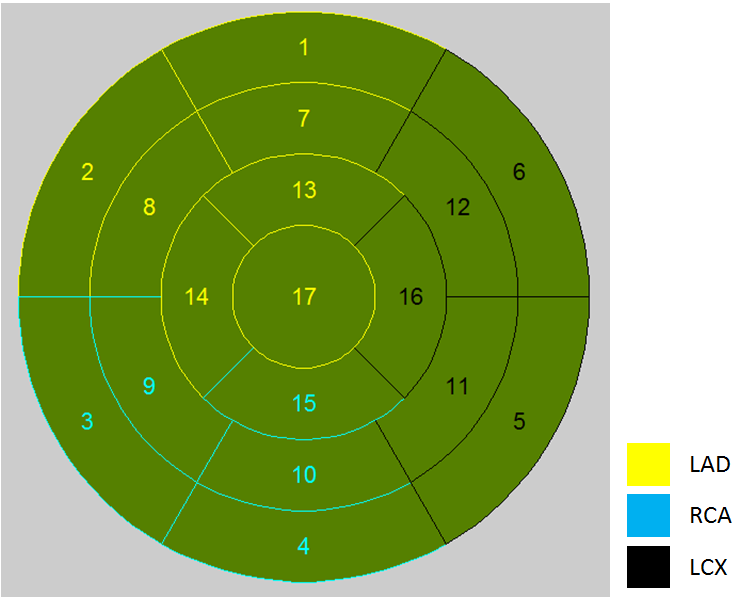}
\end{center}
   \caption{Bull's eye view of the 17-segment AHA model. The numerical label for each LV segment is color coded to denote the territory of the major coronary artery it falls under, i.e., Left Anterior Descending (LAD): yellow; Right Coronary Artery (RCA): cyan; and Left Circumflex Artery (LCX): black.}
\label{fig:17_segment_visual}
\end{figure}

The standardized myocardial segmentation model proposed by the American Heart Association (AHA) is adopted to facilitate accurate segmentation, understanding and localization of cardiac anatomy and pathology~\cite{Medtronic}. The AHA-approved 17-segment cardiac model~\cite{cerqueria02} is adapted to divide the left ventricle into 17 segments for more accurately localized shape analysis (Figure~\ref{fig:17_segment_visual}). The long axis of the left ventricle is first computed to divide the LV endocardial surface into 4 main parts along the longitudinal orientation termed as the {\it apex, apical, mid-cavity} and {\it basal}. Further division of the LV endocardial surface along the short axis view is tackled by exploiting knowledge of cardiac anatomy. Three landmark points are considered across the septum to carry out the division process. This results in the division of {\it apical} segment into 4 parts and the {\it mid-cavity} and {\it basal} segments each into 6 parts whereas the {\it apex} remains undivided. At the end of the segmentation process, the LV endocardial surface is divided into 17 segments in conformity to the AHA model as shown in Figure~\ref{fig:17_segment_visual}.

\subsection{LV endocardial surface descriptors based on the rigidity assumption} \label{subsec:LV_rigid}

In our preliminary work~\cite{mukhopadhyay11}, we considered two primary shape descriptors, to characterize the LV endocardial surface morphology, i.e., the $D2$ {\it shape descriptor}, a global shape descriptor proposed by Osada et al.~\cite{osada02} and the {\it shape index}, a local shape descriptor first introduced by Koenderink~\cite{koenderink90} and subsequently modified by Zaharia and Preteux~\cite{zaharia01}.  Both descriptors make an implicit assumption of rigidity of the endocardial surface and are invariant to scale and 3D rigid body transformation (i.e., rotation and translation) in Euclidean space.  

The $D2$ shape descriptor is a global shape signature of a 3D object and is represented by a probability distribution function. The probability distribution function is obtained via sampling of a pre-specified local shape function which measures a basic geometric property of the underlying 3D shape. In the case of the $D2$ shape descriptor, the shape function is the distance between two randomly sampled vertices on the mesh-based representation of the 3D LV surface. Osada et al.~\cite{osada02} have shown that the simple $D2$ shape descriptor, when modeled as a probability distribution function over the entire 3D object surface, can serve as an effective object signature for a variety of applications such as 3D shape-based retrieval from a 3D object database.  

The {\it shape index}, in contrast, describes the local shape of the 3D surface mesh based on the values of the principal surface curvatures computed within a local neighborhood of a given 3D surface point. The shape index is particularly effective in describing the local extrinsic geometry of the surface.  The shape index $I_p$ of a surface point $p$ is a function of the two principal surface curvatures $\kappa_1 (p)$  and $\kappa_2 (p)$  associated with point $p$ and is defined as follows:  

\begin{equation}
I_p = \dfrac{1}{2} - \dfrac{1}{\pi} \arctan \left ( \dfrac{\kappa_1(p) + \kappa_2(p)}{\kappa_1(p) - \kappa_2(p)} \right )  \label{eq:SI}
\end{equation}
where, $\kappa_1(p)  > \kappa_2(p)$.

The value of the shape index lies within the interval [0, 1] and enables surface-based representation of basic elementary shapes such as {\it convex, concave, rut, ridge} and {\it saddle}~\cite{zaharia01}. Since the LV endocardial surface is observed to exhibit both, global structure as well as detailed local structure and since there have been no previous studies detailing which of the shape properties (local or global) convey more valuable clinical information, we considered both local and global shape descriptors in our preliminary work~\cite{mukhopadhyay11}. However, the results of our preliminary work showed the shape index to be more effective than the $D2$ shape descriptor for classification of the LV endocardial surface samples into {\it diseased} and {\it non-diseased} classes. 

\subsection{LV endocardial surface descriptors based on non-rigid deformation} \label{subsec:LV_nonrigid}

BoF feature-based approaches for shape representation typically consist of two steps, i.e., feature detection and feature description. For representation of object shapes encountered in common imagery, the feature detectors and feature descriptors are designed to be invariant under affine transformation to take into account the different possible viewpoints under which a 3D object may appear in an image. For non-rigid characterization of the LV endocardial surface, the type of invariance required is different from the one required when dealing with common images since the number of surface transformations in the case of the former is much larger, including bending, changes in pose and changes in connectivity. In our case, the LV endocardial surface deformation can be constrained to lie within a pre-specified range since all the CTCA images are acquired at a relatively steady phase (75\%) within the R-R cardiac cycle. The aforementioned steady phase assumption allows the LV surface deformation to be approximated by an \textit{isometric} deformation. Consequently, the LV endocardial surface descriptors need to be robust enough to capture the important characteristics of the underlying shape while exhibiting invariance to isometric deformation. Since the LV endocardial surface is typically less rich in terms of features than common images, the task of detecting a large number of stable, repeatable and isometric deformation-invariant feature points is significantly harder. In this work, the features selected for description of the LV endocardial surface are inspired by the work of Toldo et. al.~\cite{toldo10}.

\subsubsection{Feature detection}
Although feature detection is not mandatory, it can improve computational efficiency in practice. In our case, we have chosen   dense random sampling as the feature detection strategy.  We considered as feature points, 500 randomly sampled surface points from each of the 17 LV endocardial surface segments. The underlying motivation was to ensure that the sampling is as dense as possible so that the entire spectrum of interesting points on the LV endocardial surface is considered for further analysis. In situations where availability of computational power is not a constraint, the feature selection step can be ignored and all the points on the surface can be considered for the ensuing feature description step.

\subsubsection{Feature description}

The feature descriptors are designed to ensure that both the local and global geometric properties of the surface are represented in sufficient detail. Inspired by the work of Toldo et. al.~\cite{toldo10}, four surface descriptors are considered to represent each sampled surface point. The first three descriptors represent the local geometry of the surface around a specific point whereas the fourth is a contextual descriptor which encapsulates the global properties of a specific point with respect to other points on the surface. The descriptors are detailed below:

\paragraph{Shape Index $(I_p)$}

The {\it shape index} $I_p$ of a surface point $p$, is computed using (equation (\ref{eq:SI})) as previously described. The shape index is particularly important for its ability to quantitatively characterize the local shape of a surface.

\begin{figure}[t]
\begin{center}
   \includegraphics[width=80mm]{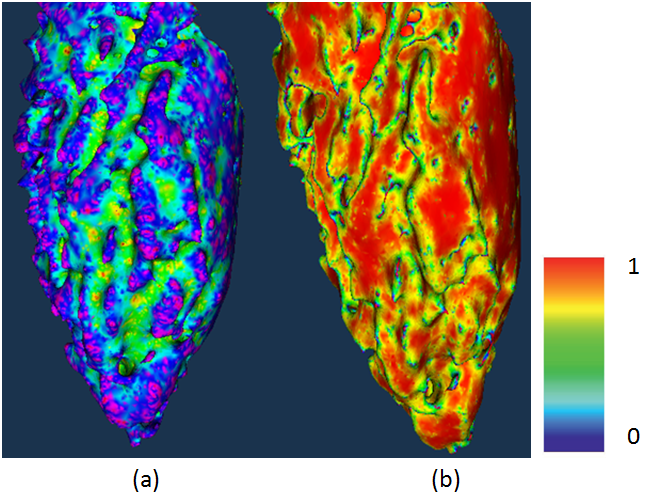}
\end{center}
   \caption{Illustration of the relationship between the (a) shape index $I_p$ and (b) curvedness $C_p$ surface descriptors for a sample LV endocardial surface.}
\label{fig:SI_curved_visual}
\end{figure}

\paragraph{Curvedness $(C_p)$}

The {\it curvedness} descriptor represents the degree of local curvature of a surface. Normalized {\it curvedness} also encapsulates the rate of change of local surface curvature: a value close to 1 implies a very gentle change, whereas a value close to 0 implies a very sharp change. Generally, points on a smooth spherical object have small {\it curvedness} values. The {\it curvedness} $C_p$ at a surface point $p$, as proposed by Koenderink~\cite{koenderink90}, is defined as a function of the two principal surface curvatures $\kappa_1 (p)$  and $\kappa_2 (p)$ associated with surface point $p$ as follows: 

\begin{equation}
C_p =  \sqrt {\dfrac{(\kappa_1 (p))^2 + (\kappa_2 (p))^2}{2} } \label{eq:CRV}
\end{equation}
where, $\kappa_1 (p)  > \kappa_2 (p)$. The relationship between the two surface descriptors, i.e., the shape index $I_p$ and the  curvedness $C_p$  for a typical LV endocardial surface is depicted in Figure~\ref{fig:SI_curved_visual}. Like the shape index $I_p$, the value of curvedness $C_p$ is also invariant to scale and 3D rigid-body transformation (i.e., translation and rotation) in Euclidean space.

\paragraph{Normal Orientation $(\theta_p)$}

It has been clinically observed that the incidence of myocardial infarction and myocardial ischemia, can cause changes in the 3D orientation of the LV trabeculation structure due to the build up of scar tissue. Consequently, the local 3D orientation of the LV endocardial surface is incorporated within the feature vector. The {\it normal orientation} $\theta_p$ at surface point $p$ is defined as the angle between the unit normal vector at surface point $p$ and the $XZ$-plane. 

%
%

\paragraph{Geodesic Contextual Descriptor $(GCD_p^{20})$}

The geodesic contextual descriptor is intended to provide a means for describing the overall shape and measuring shape similarity based on point correspondences. Contextual information is used to describe a specific surface point $p$ with respect to the 3D shape as a whole. The geodesic contextual descriptor for a surface point $p$ encapsulates the distribution of relative positions of other points on the same surface with respect to $p$, thus summarizing global shape in an informative and, most importantly, in an isometric deformation-invariant manner. 

Moreover, since BoF-based methods typically employ a frequency distribution histogram, there is inherent loss of contextual information. An example of this is a Google search based on the keyword {\it Matrix} that yields mutually distinct results; results pertaining to the movie titled {\it Matrix} and results pertaining to the concept {\it Matrix} from linear algebra. These semantically distinct results can be disambiguated only via incorporation of contextual information. Since it is difficult to incorporate contextual information within a frequency distribution histogram, it needs to be incorporated within the feature vector. The {\it geodesic contextual descriptor}, denoted by $GCD_p^{20}$, is an isometric deformation-invariant contextual descriptor that encapsulates the relative position of a surface point $p$ in relation to the other points on the surface segment. It is characterized by a 20-bin histogram which is generated by computing the normalized geodesic distance between the surface point $p$ and other sampled points on the surface segment. The geodesic contextual descriptor $GCD_p^{20}$ is invariant to scale, 3D rigid body transformation and isometric deformation.

The feature detection and feature description procedures described above result in a 23-tuple feature vector for each surface point $p$ denoted by $F_p= (I_p ,C_p , \theta_p, GCD_p^{20})$ as depicted in Figure~\ref{fig:concept_visual}.

\begin{figure}[t]
\begin{center}
 \includegraphics[width=80mm]{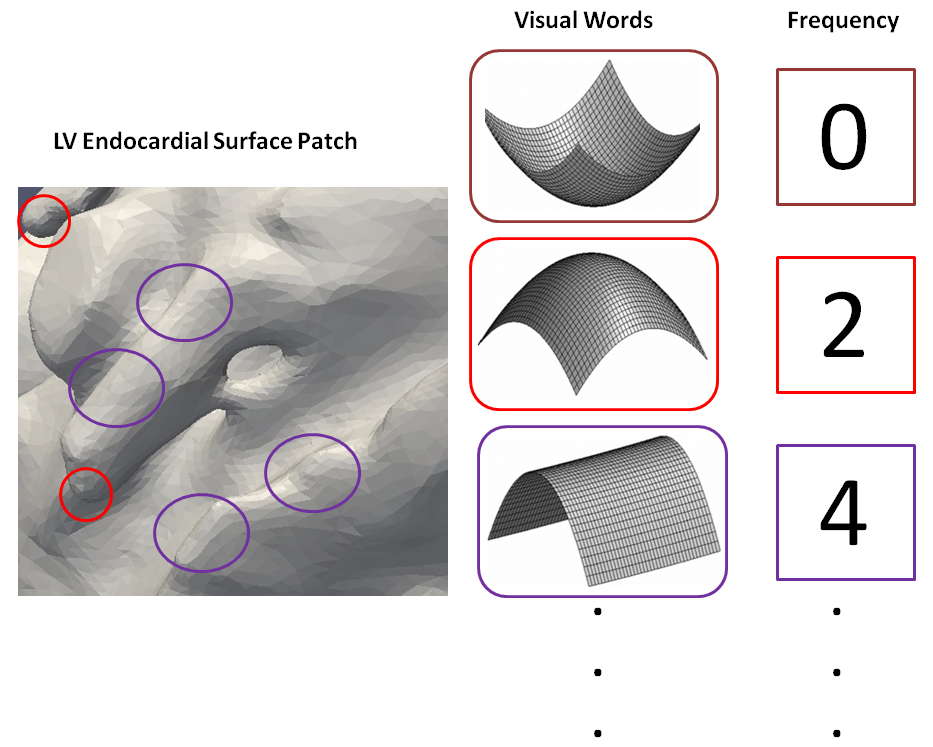}
\end{center}
   \caption{Concept visualization for the BoF histogram generation procedure. The frequency distribution histogram of the sampled {\it visual words} is illustrated for a sample LV endocardial surface.}
\label{fig:Bof_gen}
\end{figure}

\subsubsection{Construction of visual vocabularies}

The feature vectors $F_p$ at each surface point are clustered in order to obtain the {\it visual words}. Assuming that the local descriptors are computed for a set of stable surface points, the feature vector space is quantized to obtain a compact representation for the vocabulary of visual words, in a manner similar to the Shape Google approach~\cite{ovsjanikov09}. A {\it vocabulary} of visual words is defined as a set of representative vectors in the descriptor space or feature space, obtained by means of unsupervised learning, i.e., vector quantization via $k$-means clustering in our case. More formally, a {\it vocabulary} of visual words is defined as a collection $V = \{ v_1, \ldots, v_k \}$ where $v_i$ is the centroid of the $i$th cluster and the clusters represent the visual words. We have experimentally chosen $k = 20$ in the $k$-means clustering algorithm for generating the final histogram and all the subsequent experiments are based on this particular value of $k$. Figure~\ref{fig:Bof_gen} depicts the frequency distribution histogram of the sampled visual words for a sample LV endocardial surface.

\section{Experimental Results}\label{sec:result}

\subsection{Experimental setup}

The proposed methods for segmentation, meshing and endocardial surface shape description were tested on 32 MDCT data sets consisting of 16 data sets from cardiac patients and 16 data sets from normal subjects. The incidence of single- or multi-vessel obstructive CAD was found in the three major coronary arteries, i.e., the Left Anterior Descending Artery (LAD), Left Circumflex Artery (LCX) and Right Coronary Artery (RCA) using XRA, and further confirmed by Myocardial Perfusion Imaging (MPI) and Fractional Flow Reserve (FFR) tests performed on the patients. The individual territories for each of the three major coronary arteries are depicted in Figure~\ref{fig:17_segment_visual} in conformity to the  American Heart Association (AHA) convention.

The cardiac patients and normal subjects were subject to a contrast-enhanced CTCA scan on a 320-MDCT scanner using a standard protocol with electrocardiogram (ECG) gating. The resulting images were reconstructed at a relatively steady state of 75\% in the R-R cardiac cycle to ensure minimal ventricular motion. This ensures that the endocardial surfaces are reconstructed at a relatively fixed cardiac dilation stage, and the subsequent shape analysis is minimally affected by cardiac motion. This method still results in the presence of motion artifacts in the acquired images due to  various factors such as the high velocity of the ventricular motion and irregular ECG. However, the assumption of isometric deformation enables us to better account for these motion artifacts. The segmentation method described in Section~\ref{subsec:LV_seg} was used to generate topologically correct and geometrically accurate data for subsequent analysis. 

The experimental results are detailed for three different scenarios. In the first scenario, the problem is modeled as one of global classification wherein the shape index ($I_p$) histogram,  the rigid-body shape descriptor $(I_p ,C_p , \theta_p)$ histogram, the combined rigid-body and isometric deformation-invariant shape descriptor $(I_p ,C_p , \theta_p, GCD_p^{20})$ histogram and the BoF histogram are each used to classify the entire LV endocardial surface into one of two classes, i.e.,  {\it normal} and {\it diseased}. In the second scenario, the BoF-based approach is used to model the problem as one of localized classification where each LV segment is labeled as {\it normal} or {\it diseased} based on the extent of diameter stenosis (DS) in the major coronary arteries supplying blood to the LV segment under consideration. The LV segment is considered {\it normal} if the percent DS value of the corresponding major coronary artery is less than 70\% and {\it diseased} if it is $\geq$ 70\%. In the third scenario, the BoF-based approach is used to model the problem as one of multivariate regression where the exact percent DS values of each of the major coronary arteries are treated as the labels based on which the correlation coefficients for each of the LV segments are computed.

\begin{figure}[t]
\begin{center}
 \includegraphics[width=80mm]{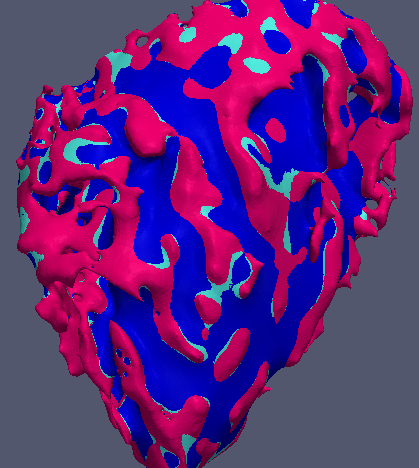}
\end{center}
   \caption{Illustration to compare the quality of the mesh generated by our segmentation algorithm. In particular, the three colors of the LV endocardial surface show the smoothness of the surface for different sizes of median filter kernel, i.e., $5\times5$ (purple), $7\times7$ (sky) and $9\times9$ (blue).}
\label{fig:segm}
\end{figure}

\subsection{Segmentation results}

The results of the LV endocardial surface segmentation have already been proven to be reasonably accurate in our preliminary study~\cite{mukhopadhyay11}. The spatial distribution of the trabeculation was observed by clinicians to vary with the location within the left ventricle; thus providing the rationale for using the standard $17$-segment AHA model to perform localized shape analysis. Furthermore, the proposed segmentation approach has already shown a visually observable distinction in trabeculation between normal and diseased hearts, yielding a classification accuracy greater than $80\%$ with simple rigid-body surface descriptors and a nearest-neighbor classifier~\cite{mukhopadhyay11}. Our previous work has demonstrated the accuracy of the proposed segmentation method as well as its applicability for subsequent quantitative shape analysis~\cite{mukhopadhyay11}. In this particular experiment, we varied the size of median filter kernel to determine its effect on the final surface mesh. The variation of the smoothness of the LV endocardial surface with respect to three different median filter kernel sizes, i.e., $5\times5$, $7\times7$ and $9\times9$ are depicted in Figure~\ref{fig:segm}. In particular, for all the later experiments, we fixed our median filter kernel size at $7\times7$ as it was observed to provide a good balance between smoothness and preservation of geometric details of the LV endocardial surface.

\subsection{Rigid shape description results}

The experimental results from our preliminary study~\cite{mukhopadhyay11} demonstrated the effectiveness and limitations of both, the $D2$ shape descriptor and shape index, for describing the LV endocardial surface. The results suggested that the information derived from the $D2$ shape descriptor is inadequate for distinguishing between normal and diseased hearts. Since the $D2$ shape descriptor is a global shape descriptor designed specifically to distinguish between objects of different classes, e.g., {\it cup} versus {\it car}, it does not capture adequately the local spatial details of the underlying shape. On the other hand, since the shape index $I_p$  is a local shape descriptor, it successfully captures the local shape details. A $20$-bin shape index histogram was generated for each LV segment. In particular, the difference in shape between the normal and diseased hearts was represented and visualized in $17\times20$ dimensions ($17$ segments per left ventricle $\times$ $1$ histogram per LV segment $\times$ $20$ bins per histogram).

\begin{figure}[t]
\begin{center}
 \includegraphics[width=80mm]{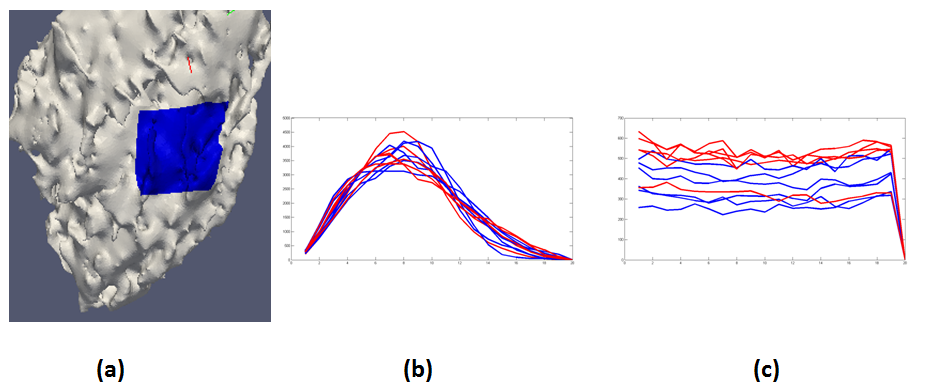}
\end{center}
   \caption{Illustration to compare the results generated by the $D2$ shape descriptor (b) and the shape index (c) for the mid-cavity segment 11 (blue) of the left ventricle (a). 12 sample datasets are considered here where the 6 normal left ventricles are represented in red and the 6 diseased ones are represented in blue in the histograms in (b) and (c).}
\label{fig:rigid_result}
\end{figure}

A typical binary classifier can be trained to separate the normal training samples from the diseased samples in high-dimensional feature space by determining a projection function that maximizes classification accuracy. Linear Discriminant Analysis (LDA) is a projection and classification method that maximizes the between-class scatter defined by $\sum_{j=1}^c (\mu_j - \mu) (\mu_j - \mu)^T $ and minimizes the within-class scatter defined by $\sum_{j=1}^c \sum_{i=1}^{N_j} (x_j^i - \mu_j) (x_j^i - \mu_j)^T $ where $\mu$  is the mean of all classes, $x_j^i$ is the $i$th sample of the $j$th class, $\mu_j$ is the mean of $j$th class, $N_j$ is the number of samples in the $j$th class and $c$ is the number of classes. The $17\times20 = 340$-dimensional feature vector generated from the Shape Index $I_p$-based histogram is projected to a subspace of $c-1$ dimensions using LDA. In our case $c=2$ since the problem is one of classification of the samples into one of two classes, i.e., {\it normal} or {\it diseased}. Classification is done using a $k$-nearest neighbor ($k$-NN) scheme in the resulting 1D subspace. 

We tested the LDA-based $k$-NN classifier for $k$ = 1 and 3, and obtained the same result: 26 out of 32 samples were classified correctly resulting in an overall accuracy of $81.25\%$. Table~\ref{tab:table1} shows the confusion matrix for the diagnosis accuracy. The false alarm rate (normal sample classified as diseased) was observed to be $18.75\%$ whereas the miss rate (diseased sample classified as normal) was observed to be $18.75\%$. The experimental results confirmed that for identification and localization of cardiovascular pathologies, local shape descriptors convey much more information than global ones. The binary classification experiment was repeated for the histograms generated by the feature vectors $(I_p, C_p, \theta_p)$ and  $(I_p ,C_p , \theta_p, GCD_p^{20})$ and the results tabulated in Tables~\ref{tab:table1_1} and \ref{tab:table2}, respectively.

\begin{table}[h]
\begin{center}
\caption{Confusion matrix to illustrate the classification accuracy of the Shape Index $I_p$-based histogram.}
\label{tab:table1}
  \begin{tabular}{| c | c | c |}
    \hline
     & Classified Diseased & Classified Normal \\ \hline\hline
    Actual Diseased & 13 &	3 \\ \hline
    Actual Normal    &  3   & 13 \\
    \hline
  \end{tabular}
\end{center}
\end{table}


\begin{table}[h]
\begin{center}
\caption{Confusion matrix to illustrate the classification accuracy of the Shape Index, Curvedness, Normal Orientation ($I_p, C_p, \theta_p$)-based histogram.}
\label{tab:table1_1}
  \begin{tabular}{| c | c | c |}
    \hline
     & Classified Diseased & Classified Normal \\ \hline\hline
    Actual Diseased & 13 &	3 \\ \hline
    Actual Normal    &  3   & 13 \\
    \hline
  \end{tabular}
\end{center}
\end{table}

The Shape Index, Curvedness, Normal Orientation ($I_p, C_p, \theta_p$)-based histogram was observed not to result in any improvement in classification over the Shape Index $I_p$-based histogram (Tables~\ref{tab:table1} and \ref{tab:table1_1}).  
In the case of the $(I_p ,C_p , \theta_p, GCD_p^{20})$-based histogram, 27 out of 32 samples were classified correctly resulting in an overall accuracy of $84.37\%$. Table~\ref{tab:table1} shows the confusion matrix for the diagnosis accuracy. The false alarm rate (normal sample classified as diseased) was observed to be $18.75\%$ whereas the miss rate (diseased sample classified as normal) was observed to be $12.5\%$. 
Thus, the incorporation of the 20-bin Geodesic Context Descriptor $GCD_p^{20})$ in the feature vector was observed to improve the classification accuracy on account of its invariance to isometric deformation. 

\begin{table}[h]
\begin{center}
\caption{Confusion matrix to illustrate the classification accuracy of the 23-tuple feature vector $(I_p, C_p, \theta_p, GCD_p^{20})$-based histogram.}
\label{tab:table2}
  \begin{tabular}{| c | c | c |}
    \hline
     & Classified Diseased & Classified Normal \\ \hline\hline
    Actual Diseased & 14 &	2 \\ \hline
    Actual Normal    &  3   & 13 \\
    \hline
  \end{tabular}
\end{center}
\end{table}


To demonstrate the superiority of the BoF-based description, the binary classification experiment described above was repeated for the BoF histogram. In this experiment, 29 out of 32 samples were classified correctly resulting in an overall accuracy of $90.62\%$. Table~\ref{tab:table3} shows the confusion matrix for the diagnosis accuracy.The false alarm rate (normal sample classified as diseased) was observed to be $6.25\%$ whereas the miss rate (diseased sample classified as normal) was observed to be $12.5\%$.

\begin{table}[h]
\begin{center}
\caption{Confusion matrix to illustrate the classification accuracy of the BoF histogram.}
\label{tab:table3}
  \begin{tabular}{| c | c | c |}
    \hline
     & Classified Diseased & Classified Normal \\ \hline\hline
    Actual Diseased & 14 &	2 \\ \hline
    Actual Normal    &  1   & 15 \\
    \hline
  \end{tabular}
\end{center}
\end{table}


\subsection{Non-rigid shape descriptor-based localized classification results} \label{subsec:non_rigid_classif}

\begin{figure}[t]
\begin{center}
 \includegraphics[width=80mm]{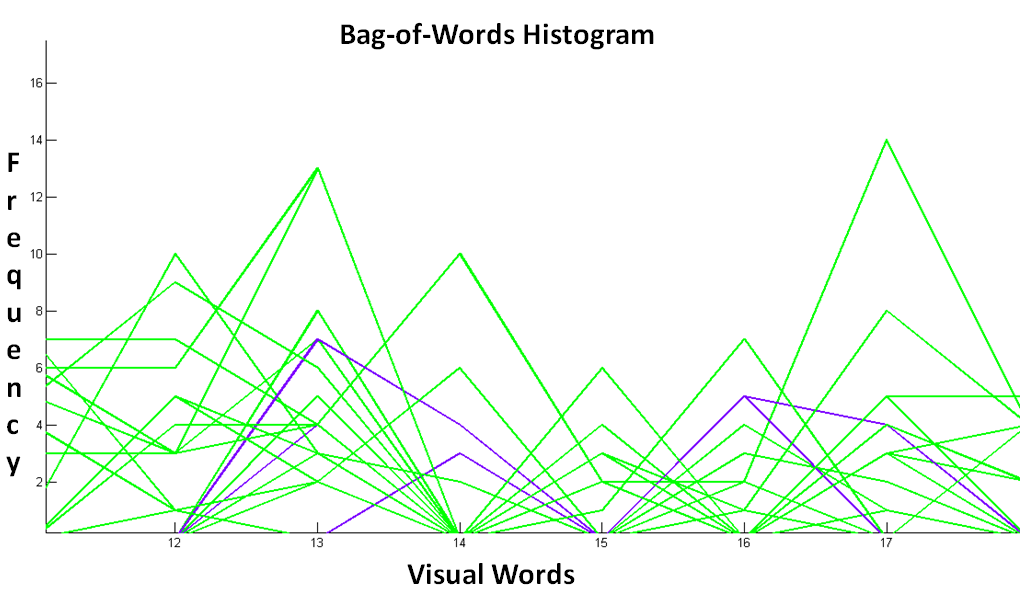}
\end{center}
   \caption{Illustration of a sample BoF frequency histogram which demonstrates its discriminative power. The normal (blue) subjects show a very distinct frequency histogram pattern when compared to the diseased subjects (green). A particular area of the BoF frequency histogram is shown here to clarify the dominant presence of diseased subjects over normal subjects in this particular region which, in turn, is reflected in the higher classification accuracy for the BoF histogram.}
\label{fig:bow_hist}
\end{figure}

A series of experiments were performed to demonstrate the effectiveness and limitations of the non-rigid shape descriptor $F_p= (I_p ,C_p , \theta_p, GCD_p^{20})$. In this series of experiments, a coronary artery was considered as {\it diseased} or {\it stenotic} if the extent of stenosis is 70\% or greater. The LV myocardial segments were labeled as diseased by a cardiologist if they were supplied blood by stenotic arteries. The already available percent Diameter Stenosis (DS) data were used for determining whether a coronary artery is {\it normal} or {\it diseased}. An artificial neural network (ANN), employing a multilayer perceptron (MLP) architecture with a single hidden layer and a learning rate of 0.3, was used for the purpose of classification of the LV segments in a manner similar to that described in~\cite{mukhopadhyay12}. The 20-bin BoF frequency histograms, generated via the vector quantization procedure (Section~\ref{subsec:LV_nonrigid}), for a particular LV segment from all the LV datasets were used as the inputs to the MLP ANN. Figure~\ref{fig:bow_hist} illustrates the discriminative power of the BoF frequency histograms in being able to distinguish between normal and diseased subjects. The ANN-based classification procedure was carried out within a strictly leave-one-out setting. The output of the MLP ANN is whether a particular LV segment can be classified as {\it normal} or {\it diseased}.

\begin{figure}[t]
\begin{center}
 \includegraphics[width=80mm]{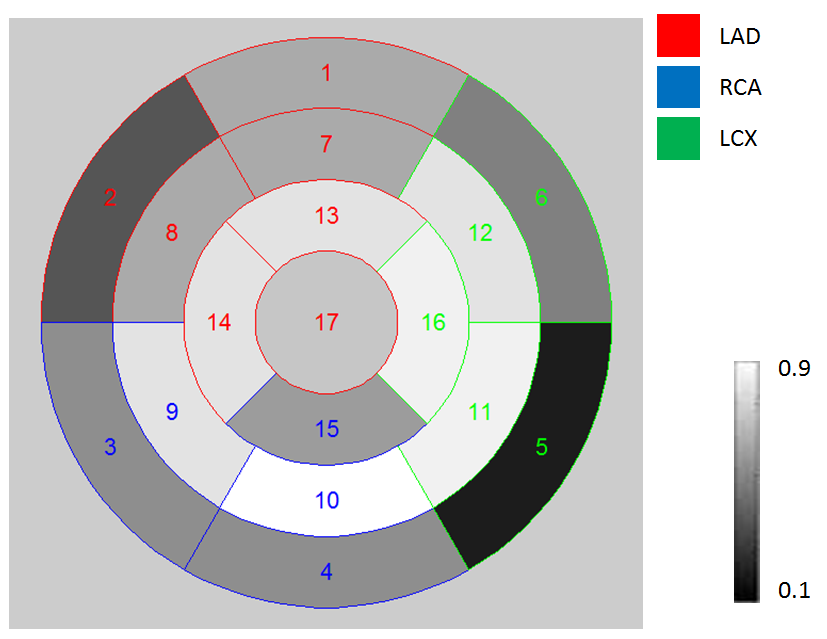}
\end{center}
   \caption{Illustration of the classification accuracy for detection of coronary arterial stenosis based on the change of surface morphology in the 17 LV segments in the AHA model. Higher gray values denotes higher classification accuracy and vice versa. The numerical label for each LV segment is color coded to denote the territory of the major coronary artery it falls under: LAD: red; RCA: blue; and LCX: green.}
\label{fig:classif_17segs}
\end{figure}

The success rate for detection of stenosis for a specific coronary artery is shown in Figure~\ref{fig:classif_17segs}. The classification results depict a clinically observed relationship between the coronary arterial stenosis and the affected segment in the 17-segment AHA model. The lower classification accuracy in the basal area (segments 1-6) can be explained by the clinical observation that several instances of coronary arterial stenosis encountered in this study are located in the mid to distal portion of the coronary arteries that only effect the mid-cavity (segments 7-12) and apical (segments 13-16) portions of the LV endocardial surface. Furthermore, another probable reason for the lower classification rate in the basal area is that the apical and mid-cavity segments exhibit greater endocardial trabeculation structure than the basal segments, which translates to more reliable endocardial surface morphology information that can be used for the purpose of classification in the case of the apical and mid-cavity segments.

\subsection{Non-rigid shape descriptor-based localized multivariate regression results} \label{subsec:non_rigid_regression}

In this experiment,  the classification problem of Section~\ref{subsec:non_rigid_classif} was recast as a multivariate linear regression problem wherein the exact percent DS data was considered as the ground truth. The regression procedure was carried out within a strictly leave-one-out setting. Multivariate linear regression attempts to fit a linear model to multiple (i.e., more than one) independent (or explanatory) variables $x_1, x_2, \ldots x_n$, to obtain an estimate for a single dependent (or response) variable $y$ as shown below: 
\begin{equation}
y =  \beta_0 + \beta_1 x_1 + \beta_2 x_2 \ldots + \beta_n x_n+\epsilon \label{eq:TrMVR}
\end{equation}
where the $\beta_i$'s are the model coefficients and $\epsilon$ is the fitting error. In our case, the dependent variable $y$ denotes the exact percent DS value and the independent variables $x_i$, $i = 1, \ldots, n=20$ denote the components of the 20-tuple BoF histogram feature vector generated via the vector quantization procedure described in Section~\ref{subsec:LV_nonrigid}. During the training process, the exact percent DS data for a major coronary artery (LAD, LCX and RCA) and the 20-tuple BoF histogram feature vectors corresponding to the LV segments that comprise the territory of that coronary artery are fed as inputs to the multivariate linear regression procedure. The multivariate linear regression procedure determines the model coefficients, i.e., $\beta_i$'s,  that minimize the mean squared error given by $\frac{1}{m} \Sigma_{k=1}^m \epsilon_m^2$ where $m$ is the number of exact percent DS value points associated with the major coronary artery (equation (\ref{eq:TrMVR})). 

During the testing procedure, the {\it estimated} DS percent value, $y_{est}$, for the major coronary is computed using the 20-tuple BoF histogram feature vector $(x_1, x_2, \ldots, x_{20})$ and the coefficients $( \beta_0, \beta_1, \ldots, \beta_{20} )$ as follows: 
\begin{equation}
y_{est} =  \beta_0 + \beta_1 x_1 + \beta_2 x_2 \ldots + \beta_n x_n \label{eq:EsMVR}
\end{equation}
The correlation coefficient is computed between the estimated outcome $y_{est}$ of equation~(\ref{eq:EsMVR}) and the exact percent DS data $y$ (i.e., the  ground truth). 


\begin{figure}[t]
\begin{center}
 \includegraphics[width=80mm]{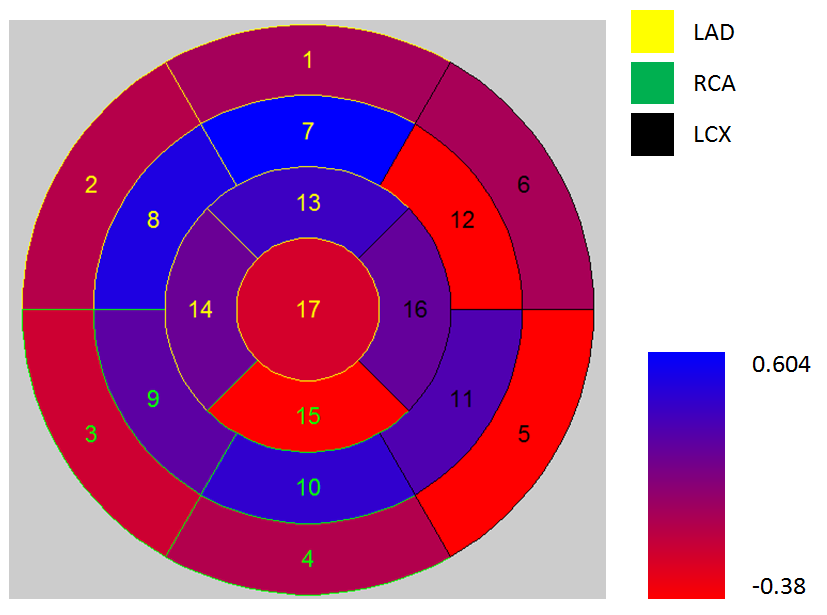}
\end{center}
   \caption{Illustration of the correlation coefficient for detection of coronary artery stenosis based on the change in surface morphology in the 17 LV segments in the AHA model. A bluish color denotes a higher correlation coefficient value whereas a reddish color denotes a lower correlation coefficient value. The numerical label for each LV segment is color coded to denote the territory of the major coronary artery it falls under: LAD: yellow; RCA: green; and LCX: black. }
\label{fig:regr_17segs}
\end{figure}

The correlation coefficient for detection of morphological changes (as quantified using the 20-tuple BoF histogram feature vectors) with respect to the percent DS values in a localized LV segment is shown in Figure~\ref{fig:regr_17segs}. The regression results were seen to depict an intriguing and clinically observed relationship between the coronary arterial stenosis and the affected LV segment in the 17-segment AHA model. The lower correlation coefficient values in the basal area (segments 1-6) compared to those in the mid-cavity (segments 7-12) and apical portions (segments 13-16) of the LV endocardial surface can be explained in a manner similar to that described in Section~\ref{subsec:non_rigid_classif}.

\section{Discussion and Conclusion} \label{sec:discussion}

To the best of our knowledge, this paper is one of the earliest works that studies the relationship between coronary artery stenosis and the morphological alterations in the LV endocardial surface using high-resolution MDCT data, and demonstrates its potential predictive value for the diagnosis of the incidence and severity of CAD. One limitation of our work is the use of the LV territories defined on the basis of the current American Heart Association (AHA) convention. However, the AHA convention may not accurately reflect the true anatomy in individual patient cases which in turn can be an explanation for the low classification accuracy in the localized classification results (Section~\ref{subsec:non_rigid_classif}).

Our investigation also sheds new light on the localization of LV regions that are the most affected by coronary artery stenosis, a phenomenon which is yet to be fully explained. The association between the morphological features of the LV endocardial surface and cardiovascular function will be explored in our future work. In particular, we aim to investigate the correlation between the LV endocardial surface morphology and the results of Myocardial Perfusion Imaging (MPI) and Fractional Flow Reserve (FFR) tests in addition to the coronary arterial stenosis results obtained via X-Ray Coronary Angiography (XRA).



\bibliographystyle{model1a-num-names}

\end{document}